\documentclass[11pt,a4paper]{article}
\usepackage[hyperref]{acl2018}
\usepackage{times}
\usepackage{latexsym}

\usepackage{url}

\usepackage{lineno}
\usepackage{multirow}
\usepackage{algpseudocode}
\usepackage{algorithm}
\usepackage{amsmath}
\usepackage{amsthm}
\usepackage{amsfonts}
\usepackage{amssymb}
\usepackage{hyperref}
\usepackage{courier}
\usepackage{enumerate}
\usepackage{graphicx}
\usepackage{helvet}
\usepackage{txfonts}
\usepackage{times}
\usepackage{titletoc}
\usepackage{tikz}
\usepackage{url}
\usepackage{setspace}
\usepackage[titletoc]{appendix}
\usepackage{subcaption}
\usepackage{url}
\usepackage{booktabs}
\usepackage{pdflscape}
\usepackage{afterpage}
\usepackage{threeparttable}

\aclfinalcopy 

\title{Temporal Analysis of Entity Relatedness and its Evolution using Wikipedia and DBpedia}

\author{Narumol Prangnawarat, John P. McCrae and Conor Hayes\\
Insight Centre for Data Analytics\\
National University of Ireland, Galway\\
IDA Business Park\\
Galway, Ireland\\
{\tt \{narumol.prangnawarat,john.mccrae,conor.hayes\}@insight-centre.org}}
\date{}

\begin{document}

\maketitle
\begin{abstract}


Many researchers have made use of the Wikipedia network for relatedness and similarity tasks. However, most approaches use only the most recent information and not historical changes in the network. 
We provide an analysis of entity relatedness using temporal graph-based approaches over different versions of the Wikipedia article link network and DBpedia, which is an open-source knowledge base extracted from Wikipedia. 
We consider creating the Wikipedia article link network as both a union and intersection of edges over multiple time points and present a novel variation of the Jaccard index to weight edges based on their transience.
We evaluate our results against the KORE dataset, which was created in 2010, and show that using the 2010 Wikipedia article link network produces the strongest result, suggesting that semantic similarity is time sensitive. We then show that integrating multiple time frames in our methods can give a better overall similarity demonstrating that temporal evolution can have an important effect on entity relatedness.





\end{abstract}


\section{Introduction}

Entity relatedness is a task that is required in many application such as entity disambiguation, recommender system and clustering. Although there are many works on entity relatedness, most of them does not consider temporal aspects. It is unclear how entity relatedness changes over time, as for example, the entities \textit{Mobile} and \textit{Camera} may have been less related in the past but may be more related at present. 
This work attempts to address this gap by showing how semantic relatedness develops over time using graph-based approaches over the Wikipedia and DBpedia~\cite{lehmann2015dbpedia,auer2007dbpedia} networks\footnote{This work focuses on the links from Wikipedia, which is reproduced as part of DBpedia graph, so these two resources are used interchangeably in this work} as well as how transient links and stable links, by which is meant links that do not persist over different times and links that persist over time respectively, affect the relatedness of the entities. We hypothesise that using graph-based approaches, such as~\cite{strube_wikirelate!_2006,hulpus_path-based_2015,leal_computing_2012}, on the Wikipedia network provides higher accuracy in term of relatedness score to the ground truth data in comparison to text-based approaches, such as~\cite{gabrilovich_computing_2007,aggarwal_wikipedia-based_2014,radinsky_word_2011}.

We show that using graph-based approaches on the Wikipedia network provides higher accuracy in term of relatedness score to the ground truth data in comparison to text-based approaches. We take each Wikipedia article as an entity, for example, the article on \textit{Semantic similarity}\footnote{\url{https://en.wikipedia.org/wiki/Semantic_similarity}} corresponds to the entity \textit{Semantic\_similarity}. Although the term \textit{article} and \textit{entity} may be referred to interchangeably in this work, \textit{article} refers to a Wikipedia article and \textit{entity} refers to a single concept in the semantic network. Wikipedia users provide Wikipedia page links within articles, so we can make use of the provided links as entities.
Assuming that entities which are closely related to an entity, a Wikipedia article in this case, are mentioned in that Wikipedia article, closely related entities share the same adjacent nodes in the Wikipedia article link network. 
However, there might be some articles that link to a lot of the articles that might not be semantically related. For example, the main page, which is the landing page for featured articles and news, is not semantically related to the articles it links to. On the other hand, articles that do not have many links to other pages might have more semantically relation to their links.  Because of this reason, we apply weights to the relationships to penalise the articles that link to many other unrelated articles.
The objective of this work is to reveal how relationships among entities affect entity relatedness as well as how temporal information affects entity relatedness. This work can be used to investigate the effect of versioning of linked data graphs, however in contrast to previous work~\cite{auer2006versioning,volkel2006semversion}, we aim to understand the trend of changes, rather than specific hierarchical changes.


%

\vspace{5mm}
\noindent
\textbf{Problem Statement and Research Contributions}

This work is extended from our previous work~\cite{prangnawarat_temporal_2017} and provides extended experimental results and a novel variation of the proposed algorithm. 
We focus on two main research problems, firstly how to represent Wikipedia network for the entity relatedness problem that varies over time. The basis of this is that Wikipedia articles have the links to other Wikipedia articles, so we can use this information to create Wikipedia article link network for the entity relatedness problem. 
A contribution of this paper towards this problem is the proposal of various temporal models to represent Wikipedia article link network to analyse entity relatedness based on the assumption that closely related entities share the same adjacent nodes in the network. We also analysed if the in-links and out-links have the same effects to their relatedness.

The second main research problem concerns method used to find relatedness and its evolution using Wikipedia network. We explore techniques for finding relatedness score of the entities. For example, there might be some articles that link to a lot of the articles that might not be semantically related, such as the main page, which is the landing page for featured articles and news. On the other hand, articles that do not have many links to other pages might have more semantically relation to their links. In this paper, we show how can we choose weights to penalise those articles that link to many other unrelated articles and how temporal information affects relatedness of entities.

We propose a graph-based approach considering entity relations, which outperforms the text-based approach in terms of the accuracy of relatedness score. We make use of the temporal Wikipedia article link network to demonstrate the evolution of entity relatedness and how transient or stable links affect the relatedness of the entities. We analyse different models of aggregated graphs, which integrate networks at various times into the same graph as well as time-varying graphs which are the series of the networks at each time step. We then evaluate this based on the KORE dataset~\cite{hoffart_kore_2012}, and show how temporal information can affect the similarity score. We show that by combining multiple versions of Wikipedia, from different time points we can produce state-of-the-art results for entity relatedness.


The remainder of this paper organised as follows: first, we discuss works related to this paper in Section~\ref{sec:relatedworks}. After that we explain how to collect the dataset from Wikipedia and DBpedia in Section~\ref{sec:dataset}. In Section~\ref{sec:methodology}, we explain methodology for representing the networks and give details of our proposed graph-based approaches to analyse entity relatedness and its evolution. We show the experiment results, which discuss the comparisons of different approaches, evolution of relatedness, relatedness with aggregated time information and frequency of the relatedness changes in Section~\ref{sec:result}.
Finally, we provide a summary of the work as well as the discussion of limitations and future works.

\section{Related Work} \label{sec:relatedworks}

This work relies on two distinct areas of research, firstly semantic relatedness that estimates the similarity between to entities and secondly temporal analysis of Wikipedia, which comprises methods for analysing change in Wikipedia. Thus, we present a brief survey of these two areas.

\subsection{Semantic Relatedness} 
Semantic relatedness works have been carried out for words (natural language texts such as common nouns and verbs) and entities (concepts in semantic networks such as companies, people and places). The approaches used for semantic relatedness includes corpus-based approaches as well as structure-based or graph-based approaches. Wikipedia and DBpedia have been widely used as resources to find semantic relatedness. However, most approaches use only the current information without temporal aspects.
One of the well-known system is WikiRelated~\cite{strube_wikirelate!_2006} introduced by Strube and Ponzetto. WikiRelated use the structure of Wikipedia links and categories to compute the relatedness between concepts.

Gabrilovich and Markovitch proposed Explicit Semantic Analysis (ESA)~\cite{gabrilovich_computing_2007} to compute semantic relatedness of natural language texts using high-dimensional vectors of concepts derived from Wikipedia.
DiSER~\cite{aggarwal_wikipedia-based_2014}, presented by Aggarwal and Buitelaar, improves ESA by using annotated entities in Wikipedia.

Leal et al.~\cite{leal_computing_2012} proposed a novel approach for computing semantic relatedness as a measure of proximity using paths on DBpedia graph. Hulpus et al.~\cite{hulpus_path-based_2015} provided a path-based semantic relatedness using DBpedia and Freebase as well as presenting the use in word and entity disambiguation.

Radinsky et al. proposed a new semantic relatedness model, Temporal Semantic Analysis (TSA)~\cite{radinsky_word_2011}. TSA computes relatedness between concepts by analysing the time series between words and finding the correlation over time. Although this work make use of the temporal aspect to find relatedness between words, it does not show how the relatedness evolves over time.

\subsection{Time-aware Wikipedia and Knowledge Bases} 
There have been a number of works in the area of time-aware Wikipedia and Knowledge Base analysis. The authors have primarily focused on content and statistical analysis.
WikiChanges \cite{nunes_wikichanges:_2008} presents a Wikipedia article’s revision timeline in real time as a web application. The application presents the number of article edits over time in daily and monthly granularity. The authors also provide the extension script for embedding a revision activity to Wikipedia.
Ceroni et. al. \cite{ceroni_information_2014} introduced a temporal aspect for capturing entity evolution in Wikipedia. They provided a time-based framework for temporal information retrieval in Wikipedia as well as statistical analysis such as the number of daily edits, Wikipedia pages having edits and top Wikipedia pages that has been changed over time.
Whiting et. al. \cite{whiting_wikipedia_2014} presented Wikipedia temporal characteristics, such as topic coverage, time expressions, temporal links, page edit frequency and page views, to find how the knowledge can be exploited in time-aware research. However, in-degree and out-degree are the only networks properties that are discussed in the paper.

Recent research from Bairi et. al.~\cite{bairi_evolution_2015} presents statistics of categories and articles, such as the number of articles, the number of links and the number of categories, comparing between the Wikipedia instance in October 2012 and the Wikipedia instance in June 2014. The authors also analysed the Wikipedia category hierarchy as a graph and provided statistics of the category graph such as number of cycles and and the cycle length of the two Wikipedia instances. 
Contropedia~\cite{borra_2015_societal} identifies when and which topics have been most controversial in Wikipedia article using Wikipedia links as the representation of the topics. However, the approach also focuses on the content changed in the article. 

A number of research also work on event analysis and extraction from Wikipedia history. 
Hienert and Luciano~\cite{hienert_extraction_2012} provided historical events extracted from Wikipedia articles as well as a timeline of extracted events. 
They parse the historical events from wikipedia using regular expressions which are language and structure specific.
Tran et. al.~\cite{tran_wikipevent:_2014} introduced a method to detect events using Local Temporal Constraint (LTC)~\cite{das_sarma_dynamic_2011} to extract complex event structures from Wikipedia.

Recent works apply time-aware knowledge bases for relations of entities. 
TimeMachine~\cite{althoff_timemachine:_2015} generates a timeline of events and relations for entities using knowledge bases. It shows the most important events or relationships that related to a given entity at the time. 
Beyond Time, recent research from Tran et. al.~\cite{Tran_Beyond_2017} introduces a novel method to compute the contextual relatedness 
using integrated time and topic models.
The temporal relatedness model uses static relatedness 
together with estimation of the dynamic relatedness by measuring a form of temporal peak coherence using Wikipedia page view statistics.
They also present the usefulness of entity relatedness with entity recommendation.
However, they do not show the evolution of the relations.

\paragraph{}
Our work makes use of the changes in Wikipedia to analyse evolution of entity relatedness. We show how entity relatedness develops over time using graph based approaches over Wikipedia network as well as how transient links and stable links affect the relatedness of the entities.

\section{Dataset Collection from Wikipedia and DBpedia} \label{sec:dataset}

\begin{figure}[!ht]
\begin{center}
\includegraphics[width=\linewidth]{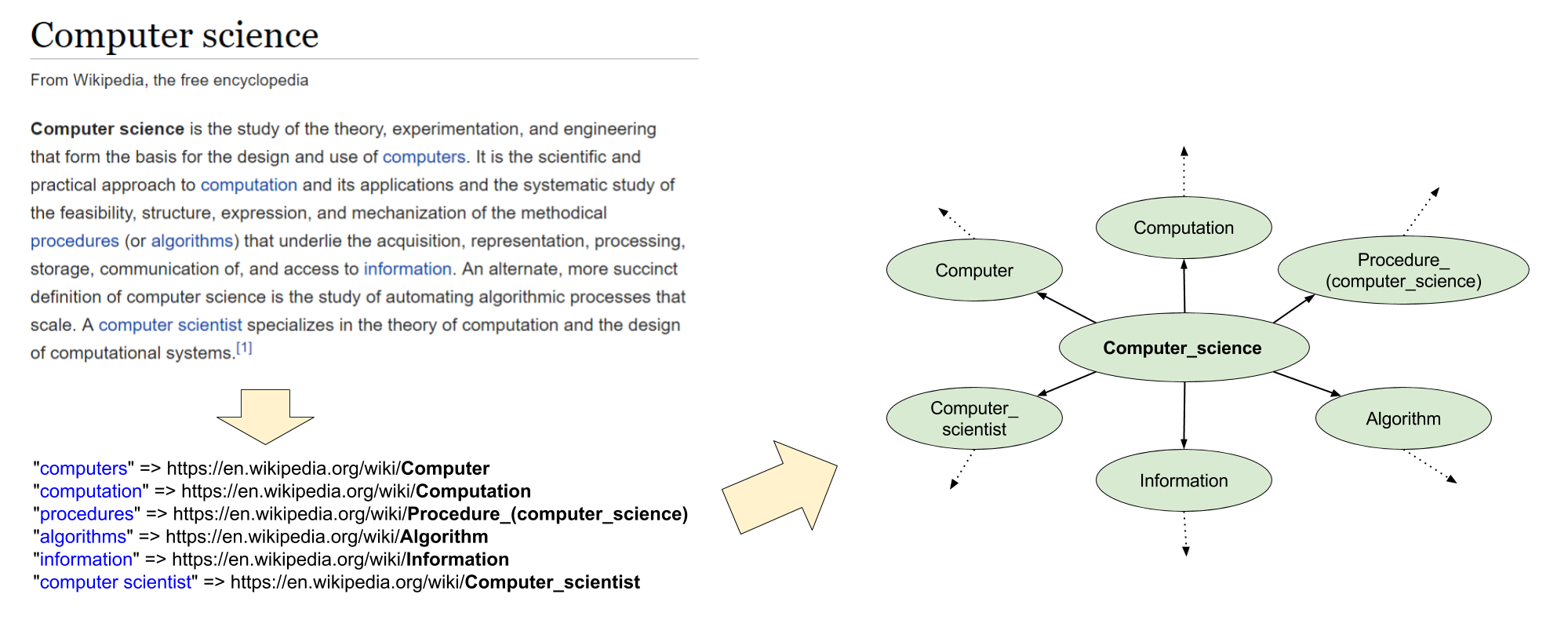}
\caption{An example of a part of Wikipedia article link network or the \textit{``Computer Science''} article}
\label{fig:wiki-network-cs}
\end{center}
\end{figure}

In this section, we explain how we collected the dataset from Wikipedia and DBpedia. DBpedia is a linked data dataset constructed from Wikipedia and we treat each Wikipedia article as an entity and a vertex in our graph. Each Wikipedia article has user provided Wikipedia article links which link to mentioned Wikipedia articles. We extract Wikipedia links within Wikipedia articles from each Wikipedia dump. Figure~\ref{fig:wiki-network-cs} shows an example of a part of the Wikipedia article link network of the \textit{``Computer Science''} article\footnote{\url{https://en.wikipedia.org/wiki/Computer_science}}. This part of the \textit{``Computer Science''} article contains links to \textit{Computer}, \textit{Computation}, \textit{Procedure\_(computer\_science)}, \textit{Algorithm}, \textit{Information}, and \textit{Computer\_scientist}. The article links of all articles in Wikipedia generate the full Wikipedia article link network.

Wikipedia articles have redirect pages which are pages that automatically redirect users to other pages. The main purpose of redirects is to redirect the article to the most appropriate article title. As Wikipedia is open, crowd-sourced information, users may use different titles to refer to the same thing such as, alternative spellings or punctuation marks, abbreviations, plurals and singulars, adjectival or adverbial forms on nouns, or likely misspellings. For instance, the article \textit{America}\footnote{\url{https://en.wikipedia.org/wiki/America}} and \textit{USA}\footnote{\url{https://en.wikipedia.org/wiki/USA}} are redirect pages that automatically redirect users to the article \textit{United States}\footnote{\url{https://en.wikipedia.org/wiki/United_States}}. 
In this case, \textit{America}, \textit{USA} and \textit{United States} should be considered as the same entity. In order to add more information to the network, we construct the Wikipedia article link network with redirects by replacing the redirect pages with the target pages 
in the Wikipedia article link network described previously. If article $a$ has a link to article $b$ and article $b$ redirects to article $c$, then we remove a link from article $a$ to article $b$ and add a link from article $a$ to article $c$. In other words, we replace a link between articles $a$ and $b$ with a link between article $a$ and article $c$.


The Wikipedia data can be downloaded from the Wikimedia download page\footnote{\url{https://dumps.wikimedia.org/}}. A major limitation of data availability from Wikipedia dumps is that the oldest available Wikipedia dump at the time of the experiment was from 20 August 2016. In order to get older Wikipedia link data, we obtained data from DBpedia\footnote{\url{http://wiki.dbpedia.org}} which is an open knowledge base extracted from Wikipedia and other Wikimedia projects. We use the page links datasets, which give the relationship of article links within each article, the same information as the Wikipedia links we extracted from Wikipedia, and redirects datasets which contains the redirects between articles in Wikipedia. Each DBpedia concept corresponds to a Wikipedia article. For example, the DBpedia concept \url{http://dbpedia.org/resource/Semantic_similarity} corresponds to the article \textit{Semantic similarity}\footnote{\url{https://en.wikipedia.org/wiki/Semantic_similarity}}. We refer to both the \textit{Semantic similarity} article and the DBpedia concept \url{http://dbpedia.org/resource/Semantic_similarity} as the entity \textit{Semantic\_similarity}. We make use of page links datasets and redirects datasets from DBpedia to construct the series of Wikipedia article link networks with redirects for each year from 2007 to 2016. The DBpedia versions used are shown in Table~\ref{tab:db-version}.

\begin{table*}[]
\centering
\caption{DBpedia versions}
\label{tab:db-version}
    \begin{tabular}{|l|l|l|}
\hline
Dataset & DBpedia version & Wikipedia Dumps \\ \hline
DBpedia 2016 & DBpedia 2016-04 & Mar-2016/Apr-2016 	\\ 
DBpedia 2015 & DBpedia 2015-04 & Feb-2015/Mar-2015 	\\ 
DBpedia 2014 & DBpedia 2014    & 02-May-2014     	\\ 
DBpedia 2013 & DBpedia 3.9     & 03-Apr-2013     	\\ 
DBpedia 2012 & DBpedia 3.8     & 01-Jun-2012     	\\ 
DBpedia 2011 & DBpedia 3.7     & 22-Jul-2011     	\\ 
DBpedia 2010 & DBpedia 3.6     & 11-Oct-2010     	\\ 
DBpedia 2009 & DBpedia 3.4     & 20-Sep-2009     	\\ 
DBpedia 2008 & DBpedia 3.2     & 08-Oct-2008     	\\ 
DBpedia 2007 & DBpedia 3.0rc   & 23-Oct-2007      \\ \hline                                         
\end{tabular}
\end{table*}

In order to examine more frequent changes, we also obtain Wikipedia data from Wikipedia dumps which are extracted from snapshots of Wikipedia pages twice a month. At the time we conducted this experiment, the newest dump available is the Wikipedia dump on 1 January 2017. We obtained and extracted Wikipedia links from 10 Wikipedia versions
as shown in Table~\ref{tab:wiki-versions}.

\begin{table}[!ht]
	\centering
	\caption{Wikipedia versions}
	\label{tab:wiki-versions}
	\begin{tabular}{|l|l|}
		\hline
		Dataset  & Wikipedia Dumps \\ \hline
		Wikipedia 2017-01-01 & 01-Jan-17       \\ 
		Wikipedia 2016-12-20 & 20-Dec-16       \\ 
		Wikipedia 2016-12-01 & 01-Dec-16       \\ 
		Wikipedia 2016-11-20 & 11-Nov-16       \\ 
		Wikipedia 2016-11-01 & 01-Nov-16       \\ 
		Wikipedia 2016-10-20 & 20-Oct-16       \\ 
		Wikipedia 2016-10-01 & 01-Oct-16       \\ 
		Wikipedia 2016-09-20 & 20-Sep-16       \\ 
		Wikipedia 2016-09-01 & 01-Sep-16       \\ 
		Wikipedia 2016-08-20 & 20-Aug-16       \\ \hline
	\end{tabular}
\end{table}

\section{Methodology} \label{sec:methodology}
We first explain various models to represent Wikipedia article link network and how we represent temporal information in the network. Then, we provide details of the extended Jaccard similarity with reciprocal centralities, our proposed graph-based approaches to analyse entity relatedness and its evolution.

\subsection{Wikipedia Article Link Network Representations}

We assume that entities which are closely related to an entity, corresponding to a Wikipedia article, are mentioned in that Wikipedia article. Hence, closely related entities share the same adjacent nodes in the Wikipedia article link network. However, there might be some articles that link to a lot of the articles that might not be semantically related. 
On the other hand, articles that do not have many links to other pages might be more semantically related to their links. 
Because of this reason, we apply weights to the relationships to penalise articles that link to many other unrelated articles. We explain more detail about the approach later in the next section.

Given Wikipedia article link information from each Wikipedia and DBpedia dataset, we construct a graph representing a snapshot of the Wikipedia information. We define both the article link network without redirect and the network with redirects, which gives more semantic information, as follows.

\begin{figure}[!ht]
\begin{center}
\includegraphics[width=\linewidth]{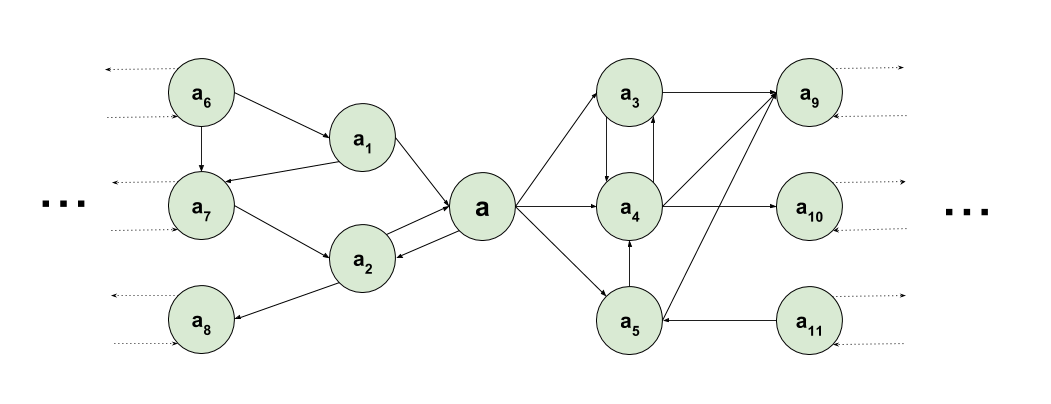}
\caption{An example of a 2-hop ego networks around the entity $a$}
\label{fig:2-hop-ego-network}
\end{center}
\end{figure}

\paragraph{Wikipedia article link network without redirect} \mbox{}\\
Given an article $a$ for a corresponding entity, a snapshot of Wikipedia article link network $G^a$ is constructed as a graph $G^a = (V^a,E^a)$ to represent a snapshot of a 2-hop ego network of article links around an entity $a$, where $V^a$ is a set of nodes where each node represents  Wikipedia entities that have links with $a$ or have links with the nodes that are adjacent to $a$ and $E^a$ is a set of edges where each edge $e_{ij}$ is an internal link between Wikipedia entities $i$ and $j$ which means an article $i$ represented by an entity $i$ mentions an entity $j$. In other words, $v \in V^a$ if $e_{av} \in E^a$ or $\exists B : e_{ab} \in E^a \wedge e_{bv} \in E^a$, i.e., $a$ is directly connected to $v$ or connected through a second node $b$. Figure~\ref{fig:2-hop-ego-network} demonstrate an example of a 2-hop ego network around the entity $a$. We construct a series of 2-hop ego networks of Wikipedia article links over time around each seed entity that we are interested in.

\begin{figure}[!ht]
\begin{center}
\includegraphics[width=\linewidth]{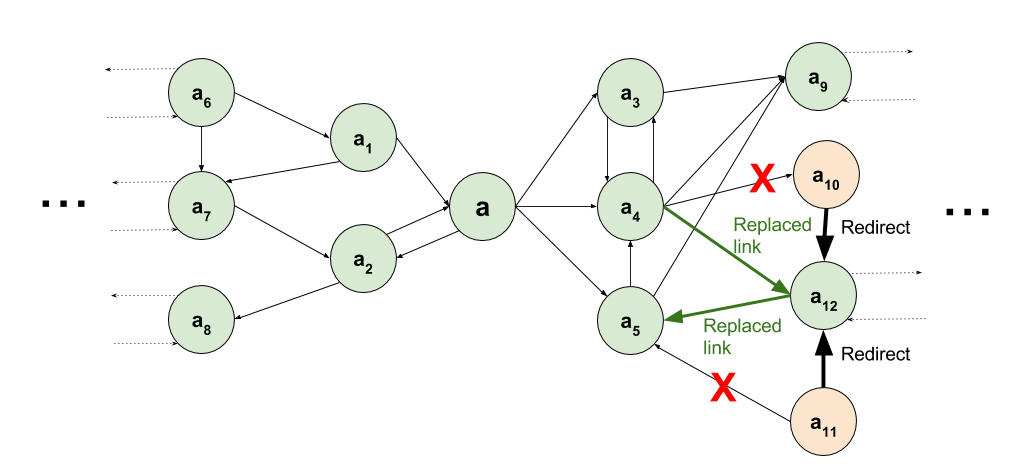}
\caption{An example of a 2-hop ego networks with redirects around the entity $a$}
\label{fig:2-hop-ego-network-redirects}
\end{center}
\end{figure}

\paragraph{Wikipedia article link network with redirects} \mbox{}\\
From the redirect information about Wikipedia, we construct graphs $G^R = (V^R, E^R)$, where $e_{ij} \in E^R$ if there is a redirect between the articles, $i$ and $j$. We use this to augment the Wikipedia article link network for an article $G^a$, by merging all nodes that have redirects. Thus, we define a \emph{Wikipedia article link network with redirects}, $G^{R,a} = (V^{R,a}, E^{R,a})$, as a graph where $e_{ab} \in E^{R,a}$ if $e_{ab} \in G^a$, $\exists i: e_{ai} \in G^a \wedge e_{ib} \in G^R$ or $\exists i: e_{ai} \in G^R \wedge e_{ib} \in G^b$.
Figure~\ref{fig:2-hop-ego-network-redirects} demonstrate an example of a 2-hop ego network with redirects around the entity $a$. We construct a series of 2-hop ego networks of Wikipedia article links with redirects over time around each seed entity that we are interested in. Entity $a_{10}$ and Entity $a_{11}$ redirect to Entity $a_{12}$ so Entity $a_{10}$ and Entity $a_{11}$ are replaced by Entity $a_{12}$ in the graph, making the link from Entity $a_4$ to Entity $a_{10}$ become Entity $a_4$ to Entity $a_{12}$ and the link from Entity $a_{11}$ to Entity $a_5$ become Entity $a_{12}$ to Entity $a_5$ instead.

\subsection{Temporal Graphs}

To reflect how relatedness changes over time, we represent temporal information from Wikipedia article link networks in two different way. One is as time-varying graphs, which are a series of Wikipedia article link snapshots at each time step. We use each version of the Wikipedia dataset and DBpedia dataset as each snapshot. Another is as an aggregated graph, which aggregates all networks at each time steps together with time information as weights. We describe each models and how we use temporal information as weights of aggregated graphs in the following sections. 

\subsubsection{Time-Varying Graphs} \label{sec:time-varying-model}

\begin{figure}[!ht]
\begin{center}
\includegraphics[width=\linewidth]{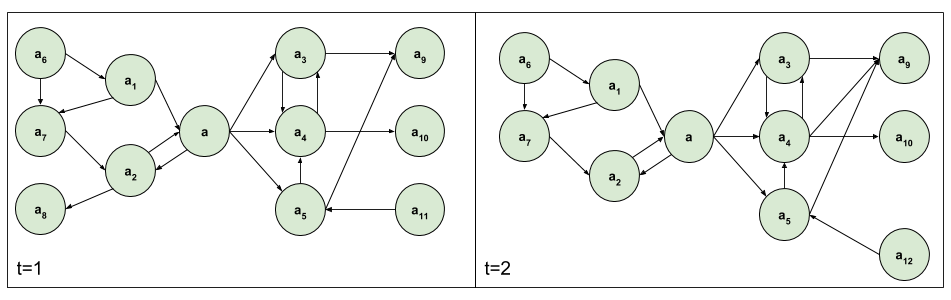}
\caption{An example of a series of time-varying graphs of an article $a$ when $T=\{1,2\}$}
\label{fig:time-varying-graphs}
\end{center}
\end{figure}

Given an article $a$ for a corresponding entity, the series of Wikipedia article link networks $G^a_S$ at the set of times $T=\{1,..,n\}$ is constructed as a set of graphs $\{G^a_1,G^a_2,...,G^a_n\}$, which we call \emph{time-varying} graphs. Each graph $G^a_t = (V^a_t,E^a_t)$ represents a snapshot of a 2-hop ego network of article links around an entity $a$ at the time $t$ as described above. 
An example of a series of time-varying graphs of an article $a$ when $T=\{1,2\}$ is shown in Figure~\ref{fig:time-varying-graphs}. We construct graphs for the link network with redirects, $G^{R,a}_t$ in the same way.

\subsubsection{Aggregated Graphs} \label{sec:aggregated-model}
In order to understand how transient links, which are links that do not persist over time, and stable links, which are links that persist over time, affect the relatedness of the entities, we create different models of aggregated graphs with differ in weights. The first model is the \textit{Intersection Model} which aggregates nodes and links that exists at all time points in the data and thus represent stable links. We also use \textit{Union Models}, which aggregate nodes and links that exists at any time points in the data and thus represent transient links. We use weights to specify how transient these linke are, including uniform temporal weight, where we weight all temporal information equally, linear temporal weight, where we weight bias toward more recent information using linear decay factor, and exponential temporal weight, where we weight bias toward more recent information using exponential decay factor. These weights are used to examine how the recency of the information affect the result of relatedness. 
We create different models of aggregated graphs for both Wikipedia article link network and Wikipedia article link network with redirects, which differ in weight described below.

\begin{figure}[!ht]
	\begin{center}
		\includegraphics[width=8cm]{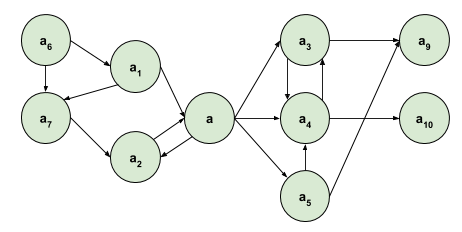}
		\caption{An example of an intersection model of an article $a$ from the previous example of the time-varying graphs}
		\label{fig:intersection-model}
	\end{center}
\end{figure}

\paragraph{Intersection model}\mbox{}\\
The intersection model aggregates nodes and links that exist in all time points in the data to represent stable links. This model captures the core of the network around each article. Nodes and edges are only included if the relationships among them exist in all time steps.  
Given a Wikipedia article $a$ for a corresponding entity, an intersection graph $G^a_I$ is constructed as a graph $G^a_I = (V^a_I,E^a_I)$, $V^a_I$ is a set of nodes where each node represents a Wikipedia article that have links with $a$ or have links with the nodes that are adjacent to $a$ at all time points and $E^a_I$ is a set of edges where each edge $e_{ij}$ is an internal link between Wikipedia entities $i$ and $j$ which appear at all time. In other words, $v \in V^a_I$ if $v \in V^a_1 \cap V^a_2 \cap ... \cap V^a_n$ and $e_{ij} \in E^a_I$ if $e \in E^a_1 \cap E^a_2 \cap ... \cap E^a_n$ for $T=\{1,..,n\}$. As this is an intersection of all time points, all edges have equal weights.
An example of an intersection model of an article $a$ from the previous example of the time-varying graphs is shown in Figure~\ref{fig:intersection-model}.

\begin{figure}[!ht]
	\begin{center}
		\includegraphics[width=8cm]{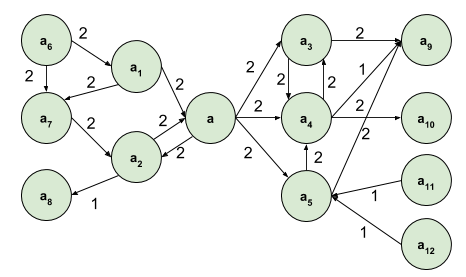}
		\caption{An example of a union model with uniform temporal weight of an article $a$ from the previous example of the time-varying graphs}
		\label{fig:union-model-uniform-weight}
	\end{center}
\end{figure}

\paragraph{Union model with uniform temporal weight} \mbox{}\\
The union model with uniform temporal weight aggregates nodes and links that exist in any time points in the data to represent transient links using uniform temporal weight, which is proportional to the number of time points containing this edge.
Given a Wikipedia article $a$ for a corresponding entity, a union graph with uniform temporal weight $G^a_U$ is constructed as a graph $G^a_U = (V^a_U,E^a_U)$, $V^a_U$ a set of nodes where each node represents a Wikipedia article that have links with $a$ or have links with the nodes that are adjacent to $a$ at any time in $T=\{1,..,n\}$ and $E^a_U$ is a set of edges where each edge $e_{ij}$ is an internal link between Wikipedia entities $i$ and $j$ with a weight $w_{ij}$ representing how many time steps the relationship appears at any time in $T$. In other words, $v \in V^a_U$ if $v \in V^a_1 \cup V^a_2 \cup ... \cup V^a_n$ and $e_{ij} \in E^a_U$ if $e_{ij} \in E^a_1 \cup E^a_2 \cup ... \cup E^a_n$ for $T=\{1,..,n\}$. The weight $w_{ij}$ of $e_{ij}$ is $a_{ij1} + ... + a_{ijn}$ for $T=\{1,..,n\}$ where $a_{ijt}=1$ if there exist the link between $i$ and $j$ at time $t$, otherwise, $a_{ijt}=0$.
An example of a union model with uniform temporal weight of an article $a$ from the previous example of the time-varying graphs is shown in Figure~\ref{fig:union-model-uniform-weight}. The numbers represent temporal weight of the relations.

\paragraph{Union model with linear temporal weight} \mbox{}\\
The union model with linear temporal weight is such that the information over different time points declines at a constant rate so the temporal factor is calculated using the linear decay factor. This weighting toward more recent links to understand how recent information affect the result of relatedness. 
Where $n$ is the newest time steps, $dt$ is the time step different, and $r$ is decay factor, non-negative temporal factor $\alpha$ can be calculate as $\alpha_{n-dt}=\alpha_n-r \cdot dt$. The newest data has 100\% temporal factor. As we have 10 time steps, we simply choose decay factor $r$ as 10\%. 
Given a Wikipedia article $a$ for a corresponding entity, $a$, $G^a_T = (V^a_T,E^a_T)$, $V^a_T$ is a set of nodes where each node represents a Wikipedia article that have links with $a$ or have links with the nodes that are adjacent to $a$ at any time in $T$ and $E^a_T$ is a set of edges, where each edge $e_{ij}$ is an internal link between Wikipedia entities $i$ and $j$ with a weight $w_{ij}$ representing summary of temporal factor where the relation appear at that time. The newer data has the higher temporal factor than the older data. The weight $w_{ij}$ of $e_{ij}$ is $(\alpha_1\cdot a_{ij1} + ... + \alpha_n\cdot a_{ijn})$ for $T=\{1,..,n\}$ where $\alpha_t$ is the temporal factor at time $t$ and $a_{ijt}=1$ if there exist the link between $i$ and $j$ at time $t$, otherwise, $a_{ijt}=0$.

\paragraph{Union model with exponential temporal weight} \mbox{}\\
The union model with exponential temporal weight has the same idea as the union model with linear temporal weight but the temporal factor is calculated using the exponential decay factor instead of the linear decay factor.
Where $n$ is the newest time steps, and $dt$ is the time step different, 
non-negative temporal factor $\alpha_e$ can be calculate as $\alpha_{n-dt}=\alpha_n\times(e^{-dt})$. The newest data has 100\% temporal factor. 

\subsection{Extended Jaccard Similarity with Reciprocal Centralities}
The Jaccard similarity coefficient measures similarity between two objects using binary attributes, which has been shown to be a robust method for semantic similarity and clustering on Web-scale graphs~\cite{strehl2000impact}. Given objects $a$ and $b$ with a vector of features $A$ and $B$ respectively, the Jaccard similarity coefficient of $a$ and $b$, $J(a,b)$ is computed by the following equation:

$$J(a,b) = \frac{|A \cap B|}{|A \cup B|}$$

Taking adjacent nodes of an entity $a$ in the Wikipedia article link network as the features of the entity $a$, the Jaccard similarity coefficient can reflect our assumption that closely related entities share the same adjacent links in Wikipedia article links network. However, the Jaccard similarity coefficient cannot take into account non-binary features. As we discussed before, there might be some articles that link to a lot of the pages that might not be semantically related so we want to apply weights to the relationships to penalise pages that have relations with many unrelated pages.
An article that is mentioned in a lot of articles may just be a general article that is not semantically related to them. 
On the other hand, an article that mentions a lot of other articles may not be semantically related to the articles that they link to. 
Centrality measures are used to rank the importance of nodes in the network, and in this work we use the following measures

\begin{description}
\item[Degree centrality:] This is a measure which counts edges connecting to the nodes. All neighbours of the node are equivalent in degree centrality.
\item[PageRank:] PageRank~\cite{page_pagerank_1999} is a centrality measure that is used to rank the importance of nodes. Neighbours of the node are not counted equivalently but weights are given to the neighbours from the idea that a node is more important if it receives more links from other nodes.
It was originally created to rank web pages in the World Wide Web network in Google search engine. We use the same idea applied to our Wikipedia article link network. 
\end{description}

To measure the similarity of nodes, we use Tanimoto's generalization~\cite{tanimoto_elementary_1958} of Jaccard similarity, which defines the similarity in terms of bit vectors as
\[
T(a,b) = \frac{\mathbf{a}^T\mathbf{b}}{||\mathbf{a}||^2 + ||\mathbf{b}||^2 - \mathbf{a}^T\mathbf{b}}
\]
We can use this formula weighted by reciprocal degree centrality and reciprocal PageRank, which are 1 divided by degree centrality and 1 divided by PageRank score respectively, to find similarity between each entity in the network. 
The underlying assumption is that articles with lower centrality scores might have more semantically relation to their links as they only link to fewer articles that are really related to them. 
For the union graph models, in addition to the reciprocal centrality scores, we also use the graph weight, which is the temporal information described above, with reciprocal centrality scores as the vector scores to compute the relatedness.




Given $\mathbf{d}_a$ is a vector of reciprocal degree centrality or PageRank scores of the articles having links with an entity $a$ in the 2-hop ego network around an entity $a$ and $\mathbf{d}_b$ is a vector of reciprocal degree centrality or PageRank scores of the articles having links with an entity $b$ in the 2-hop ego network around an entity $b$. The vector value is zero for unconnected articles. The relatedness between two entities $a$ and $b$ using Extended Jaccard Similarity with Reciprocal Centrality is computed as:

$$R(a,b) = \frac{\mathbf{d}_a^T \mathbf{d}_b}{||\mathbf{d}_a||^{2}+||\mathbf{d}_b||^{2}-\mathbf{d}_a^T \mathbf{d}_b}$$

Given $\mathbf{tw}_a$ is a vector of temporal weight between entity $a$ and other nodes in graphs and $\mathbf{tw}_b$ is a vector of temporal weight between entity $b$ and other nodes in graphs. the relatedness between two entities $a$ and $b$ using Extended Jaccard Similarity with Temporal Weight and Reciprocal Centrality is computed as\footnote{$\circ$ denotes the Hadamard product (element-wise multiplication)}:

\begin{eqnarray*}
    R_{TW}(a,b) =& \frac{(\mathbf{d}_a \circ \mathbf{tw}_a)^T (\mathbf{d}_b \circ
    \mathbf{tw}_b)}{x_{TW}}\\
    x_{TW} =& ||(\mathbf{d}_a \circ \mathbf{tw}_a)||^{2}+\\&||(\mathbf{d}_b \circ
    \mathbf{tw}_b)||^{2}-\\&(\mathbf{d}_a \circ \mathbf{tw}_a)^T (\mathbf{d}_b \circ \mathbf{tw}_b)
\end{eqnarray*}

\section{Experiment Results} \label{sec:result}


We conducted experiments and evaluated with the KORE~\cite{hoffart_kore_2012} dataset in order to show the effectiveness of our methods for temporal analysis of entity relatedness and its evolution. The KORE dataset has been created to measure relatedness between named entities. It consists of 420 related entity pairs from a selected set of 21 seed entities from the YAGO2~\cite{hoffart_yago2_2013} knowledge base from 4 different domains, which are 5 entities from IT companies, 5 entities from Hollywood celebrities, 5 entities from video games, 5 entities from television series, and one singleton entity. Each of the entities has 20 ranked related entities. All entities in the KORE dataset corresponds to entities in the Wikipedia article link networks. The KORE dataset was chosen as it is one of the most widely-used and is sufficiently large to make statistical significant results. Moreover, it was created several years ago in 2010, thus meaning that there are at least 5 years of data from Wikipedia available both before and after its creation, allowing us to examine the effect of temporal change more effectively.

We use Spearman Correlation to compare the relatedness scores from each approach with the scores from the KORE dataset. As the KORE dataset provides only the ranking but not the score, we assume that the highest entity has a score of 20 and each subsequent entity has a score 1 lower.

\subsection{Extended Jaccard Similarity with Reciprocal Centralities Results}


For each DBpedia version stated above, we constructed a series of networks of entities, as described in Section~\ref{sec:time-varying-model}, seeding the entities from KORE dataset. 

The experiments were conducted to compare three different perspectives. In the first evaluation perspective, we performed experiments to show that the proposed graph-based extended Jaccard similarity with reciprocal centralities outperforms the text-based approach and the baseline graph-based approaches in terms of the relatedness score accuracy. 
We used Term Frequency-Inverse Document Frequency (TF-IDF) based similarity as the text-base baseline, and binary Jaccard similarity as the baseline for the graph-based approach to compare to the proposed approaches. 
TF-IDF was calculated using \texttt{scikit-learn}'s\footnote{\url{http://scikit-learn.org}} functions, treating each Wikipedia article as a single document for the document frequency and counting the term frequency of each word, without stop words, in the article corresponding to the entities to be compared.
We analysed variations of features for binary Jaccard methods by considering only in-links of the nodes which are the entities that have links to the nodes, only out-links of the nodes which are the entities that have links from the nodes, and both in-links and out-links of the nodes which are entities that have links to or from the nodes.The second evaluation perspective aims to show different results between using reciprocal degree centrality and reciprocal PageRank in the proposed similarity methods. And the third perspective is to show that the Wikipedia article link network with redirects give better result than the Wikipedia article link network without redirect. 

We performed TF-IDF based similarity on three different Wikipedia text revisions. One is the revision at the time when YAGO2 was created (17-Aug-2010) which is used to constructed the KORE dataset. The second one is the revision at the dump time of DBpedia 2009 dataset (20-Sep-2009) and the last one is the revision at the dump time of DBpedia 2010 dataset (11-Oct-2010). We performed Spearman Correlation to evaluate our system's correlation with the gold standard dataset, KORE dataset, as described previously. The Spearman Correlations of the 3 different datasets compared to the KORE gold standard are shown in Table~\ref{tab:bag-of-word-results}. We can see that the result of the data acquired at the time when YAGO2 was created has the highest correlation as the same information is captured at that time. It is significantly better than the result of Wikipedia at the time of DBpedia 2009 dump with p-value $<$ 0.1 and slightly better than the result of Wikipedia at the time of DBpedia 2010 dump.

\begin{table}
\centering
\caption{Spearman Correlations to the KORE gold standard with the text-based approach for different Wikipedia dumps}
\label{tab:bag-of-word-results}
    \begin{tabular}{|p{5cm}|l|}
\hline
\textbf{Version}               & \textbf{Correlation} \\ \hline
Wikipedia at the time when YAGO2 was created                & 0.503212             \\ \hline
Wikipedia at the time of DBpedia 2009 dump & 0.491440              \\ \hline
Wikipedia at the time of DBpedia 2010 dump & 0.502987             \\ \hline
\end{tabular}
\end{table}

\begin{table*}
\centering
\captionof{table}{Spearman Correlations to the KORE gold standard comparing text-based approach with graph-based approaches between models without redirects and models with redirects of data from DBpedia 2009 and DBpedia 2010}
\label{tab:corr-2009-2010-redirects}
\begin{threeparttable}
    \begin{tabular}{|p{40mm}|l|l|l|l|}
\hline
\multirow{2}{*}{}             & \multicolumn{2}{c|}{Dbpedia 2009}                                           & \multicolumn{2}{c|}{Dbpedia 2010}                                           \\ \cline{2-5} 
                              & \multicolumn{1}{c|}{Without Redirect} & \multicolumn{1}{c|}{With Redirects} & \multicolumn{1}{c|}{Without Redirect} & \multicolumn{1}{c|}{With Redirects} \\ \hline
TF-IDF                        & 0.491440 $\ast\ast$                         & -                                   & 0.502987 $\ast\ast\ast$                        & -                                   \\ \hline
Jaccard (I)                   & 0.568509 $\star$                          & 0.575758 $\ast\ast$                       & 0.568026 $\ast\ast$                         & 0.588351 $\ast\ast$                       \\ \hline
Jaccard (O)                   & 0.511564 $\ast$                          & 0.504138 $\ast\ast\ast$                      & 0.559898 $\ast\ast$                         & 0.564088 $\ast\ast\ast$                      \\ \hline
Jaccard (I+O)                 & 0.578706 $\star$                          & 0.580944 $\ast\ast$                       & 0.585535 $\ast\ast$                         & 0.586593 $\ast\ast\ast$                      \\ \hline
Extended Jaccard RP (I)            & 0.585212 $\star$                          & 0.591538 $\ast\ast$                       & 0.569857 $\ast\ast$                         & 0.616843 $\ast\ast$                       \\ \hline
Extended Jaccard RP (O)            & 0.578478 $\star$                          & 0.566556 $\ast\ast\ast$                      & 0.576616 $\ast\ast\ast$                        & 0.601197 $\ast\ast\ast$                      \\ \hline
Extended Jaccard RP (I+O)          & 0.604112                              & 0.613450 $\ast\ast$                       & 0.599284 $\ast\ast$                         & 0.637930 $\ast\ast$                       \\ \hline
Extended Jaccard RD (I)            & 0.623768                              & 0.634961 $\star$                        & 0.632760 $\ast$                          & 0.672512 $\star$                        \\ \hline
Extended Jaccard RD (O)            & 0.588592 $\star$                          & 0.567652 $\ast\ast\ast$                      & 0.601755 $\ast\ast$                         & 0.609715 $\ast\ast\ast$                      \\ \hline
\textbf{Extended Jaccard RD (I+O)} & 0.639055                              & \textbf{0.651450}                   & 0.658647 $\star$                          & \textbf{0.696506}                   \\ \hline
\end{tabular}
\begin{tablenotes}[para,flushleft]
	$\star$, $\ast$, $\ast\ast$, and $\ast\ast\ast$ mean the Wikipedia article link network with redirects using the Extended Jaccard with Reciprocal Degree Centrality considering both in-links and out-links is significantly better than this result with p-value $<$ 0.1, p-value $<$ 0.05, p-value $<$ 0.01 and p-value $<$ 0.001 respectively.
\end{tablenotes}
\end{threeparttable}
\end{table*}

We compared the text-based approach of each version of dataset to the graph-based approaches. In this section, we focused on DBpedia 2009 dataset and DBpedia 2010 dataset as they are the closest snapshots to the Wikipedia dump from 2010-08-17 which is used to constructed YAGO2 using by the KORE dataset. We found that the graph-based approaches significantly outperform the result from the text-based approach in term of accuracy of relatedness score as shown in Table~\ref{tab:corr-2009-2010-redirects}. Moreover, the results show that the proposed Extended Jaccard Similarity with reciprocal centralities give a better accuracy of relatedness score than the baseline binary Jaccard methods. The extended Jaccard similarity with reciprocal degree centrality gives a better accuracy of relatedness score than the extended Jaccard similarity with reciprocal PageRank. The extended Jaccard similarity with reciprocal degree centrality by considering both in-links and out-links gives the best result in terms of accuracy of relatedness score.
This shows that considering the number of entity links improves the relatedness accuracy but not the weights from the neighbours.

Moreover, after we add more information to the graph by applying redirect relations to the model, the correlation results improve. Table~\ref{tab:corr-2009-2010-redirects} shows the Spearman Correlations to the KORE gold standard comparing text-based approach with graph-based approaches between models without redirects and models with redirects of data from DBpedia 2009 and DBpedia 2010. We use I for in-links, O for out-links, I+O for both in-links and out-links, RD for reciprocal degree centrality and RP for reciprocal PageRank. For example, Ext Jaccard RD (I+O) means the extended Jaccard similarity with reciprocal degree centrality by considering both in-links and out-links. We can see that using the extended Jaccard similarity with reciprocal degree centrality by considering both in-links and out-links with the Wikipedia article link network with redirects gives the best result. $\star$,$\ast$,$\ast\ast$ and $\ast\ast\ast$ show that our best performing method (Extended Jaccard with Reciprocal Degree Centrality considering both in-links and out-links) is significantly better than this result with p-value $<$ 0.1, p-value $<$ 0.05, p-value $<$ 0.01 and p-value $<$ 0.001 respectively for each star symbol.

\subsection{Evolution of Relatedness} 
\begin{figure}[h]
\begin{center}
\includegraphics[width=\linewidth]{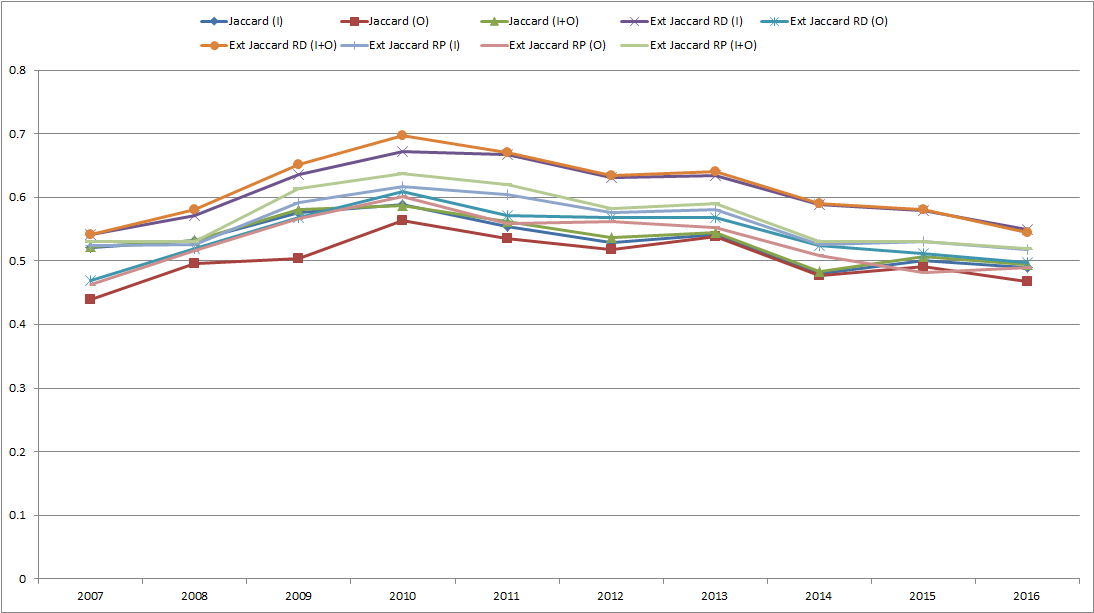}
\caption{The comparison of Spearman Correlations to the KORE gold standard of different graph-based methods from each DBpedia dataset with redirects}
\label{fig:corr-chart-redirects}
\end{center}
\end{figure}

We analysed the change of correlations to the KORE gold standard in different datasets to demonstrate how relatedness progresses over subsequent years. We found that the dataset from the network in 2010 got the highest results. This is because the KORE dataset is constructed using YAGO2 that use the Wikipedia dump from 2010-08-17~\cite{hoffart_yago2_2013}, which is the closest to the DBpedia version 2010 (2010-10-11). Figure~\ref{fig:corr-chart-redirects} shows the comparison of Spearman Correlation result of different methods from each dataset with redirects.

\begin{figure}[ht]
	\begin{center}
		\includegraphics[width=\columnwidth]{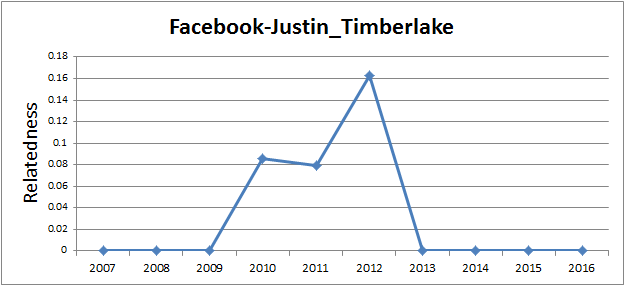}
		\caption{ Evolution of relatedness between \textit{Facebook} and \textit{Justin\_Timberlake}}
		\label{fig:facebook}
	\end{center}
\end{figure}

We can see from the result that the most up-to-date knowledge bases will not give the highest correlation if the ground truth data is created in different time. This is because the relatedness score varies according to the relatedness of the entities at that time. For instance, \textit{Facebook} and \textit{Justin\_Timberlake} has a higher relatedness score in 2010 as \textit{Justin\_Timberlake} stared in \textit{The Social Network} movie, the film portrays the founding of \textit{Facebook}, and the score faded after that as shown in Figure~\ref{fig:facebook}. 
The relatedness scores are computed using the extended Jaccard similarity with reciprocal degree centrality which gives the best result as shown in the previous section.

\begin{figure}[ht]
\begin{center}
\includegraphics[width=\columnwidth]{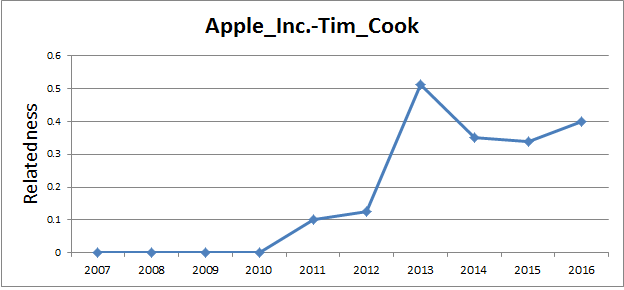}
\caption{Evolution of relatedness between \textit{Apple\_Inc.} and \textit{Tim\_Cook}}
\label{fig:apple}
\end{center}
\end{figure}

We also did a qualitative analysis for the entities which are not in the KORE dataset as they become more related to the seed entities after the dataset was created. For instance, \textit{Tim\_Cook} started to have high relatedness with \textit{Apple\_Inc.} since 2011 when he become the CEO of the company. Figure~\ref{fig:apple} shows the relatedness of \textit{Apple\_Inc.} and \textit{Tim\_Cook}.

\begin{figure}[ht]
\begin{center}
\includegraphics[width=\columnwidth]{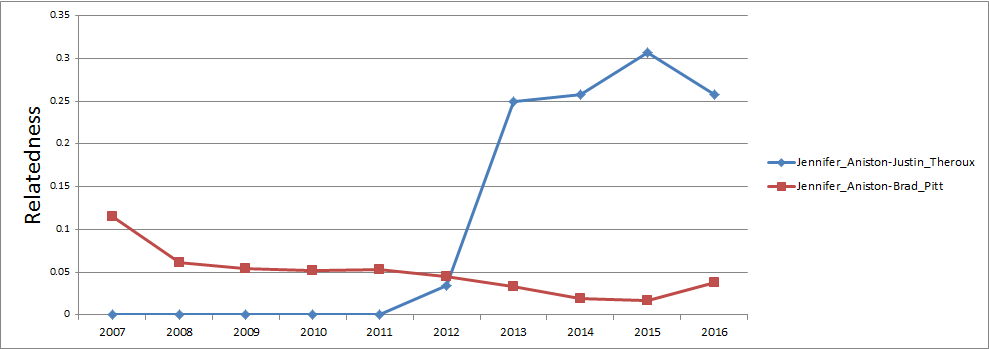}
\caption{Evolution of relatedness between \textit{Jennifer\_Aniston} and \textit{Justin\_Theroux} in comparison to \textit{Jennifer\_Aniston} and \textit{Brad\_Pitt}}
\label{fig:jennifer}
\end{center}
\end{figure}

Another example is the relatedness between \textit{Jennifer\_Aniston} and \textit{Justin\_\-Theroux}\footnote{\url{https://en.wikipedia.org/wiki/Justin_Theroux}} which became higher from 2011 when they started dating\footnote{\url{http://people.com/celebrity/jennifer-aniston-justin-theroux-engaged/}}. While the relatedness between \textit{Jennifer\_Aniston} and \textit{Brad\_Pitt} stayed high for a while after they divorced in 2005\footnote{\url{http://people.com/celebrity/week-ahead-brad-jen-finalize-divorce/}} and then faded as shown in Figure~\ref{fig:jennifer}.

\begin{figure}[!ht]
	\begin{center}
		\includegraphics[width=\columnwidth]{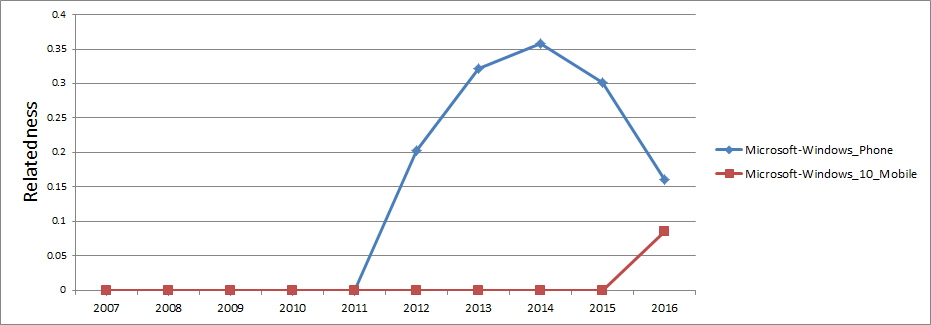}
		\caption{Evolution of relatedness between \textit{Microsoft} and \textit{Windows\_Phone} in comparison to \textit{Microsoft} and \textit{Windows\_10\_Mobile}}
		\label{fig:microsoft}
	\end{center}
\end{figure}

Figure~\ref{fig:microsoft} shows the comparison of the evolution of relatedness between \textit{Microsoft} and \textit{Windows\_Phone} and between \textit{Microsoft} and \textit{Windows\_10\_Mobile}. We can see that \textit{Windows\_Phone} started to have relatedness with \textit{Microsoft} after the first initial release in 2010 and the score faded after it was replaced by \textit{Windows\_10\_Mobile} which is first announced in 2015\footnote{\url{https://en.wikipedia.org/wiki/Windows_Phone}}\textsuperscript{,}\footnote{\url{https://en.wikipedia.org/wiki/Windows_10_Mobile}}. 

We performed the Spearman Correlation between each dataset and found that the correlations of the entity relatedness are higher to the dataset versions that are closer in time to themselves and lower when the time is more different as shown in Table~\ref{tab:corr-diff}. This shows that the entity relatedness gradually changes over time.

		\begin{table*}[]
		\centering
		\captionof{table}{Correlations of the entity relatedness between each DBpedia dataset with redirects}
\label{tab:corr-diff}
\begin{tabular}{|l|l|l|l|l|l|l|l|l|l|}
\hline
\textbf{Dataset} & \textbf{2008} & \textbf{2009} & \textbf{2010} & \textbf{2011} & \textbf{2012} & \textbf{2013} & \textbf{2014} & \textbf{2015} & \textbf{2016} \\ \hline
\textbf{2007}    & 0.717         & 0.635         & 0.611         & 0.605         & 0.579         & 0.581         & 0.513         & 0.512         & 0.500         \\ \hline
\textbf{2008}    &               & 0.785         & 0.724         & 0.723         & 0.704         & 0.699         & 0.633         & 0.605         & 0.588         \\ \hline
\textbf{2009}    &               &               & 0.843         & 0.818         & 0.773         & 0.772         & 0.703         & 0.678         & 0.627         \\ \hline
\textbf{2010}    &               &               &               & 0.855         & 0.805         & 0.807         & 0.728         & 0.701         & 0.654         \\ \hline
\textbf{2011}    &               &               &               &               & 0.864         & 0.859         & 0.779         & 0.740         & 0.701         \\ \hline
\textbf{2012}    &               &               &               &               &               & 0.931         & 0.845         & 0.805         & 0.757         \\ \hline
\textbf{2013}    &               &               &               &               &               &               & 0.862         & 0.819         & 0.765         \\ \hline
\textbf{2014}    &               &               &               &               &               &               &               & 0.894         & 0.838         \\ \hline
\textbf{2015}    &               &               &               &               &               &               &               &               & 0.893         \\ \hline
\end{tabular}
\end{table*}

\subsection{Relatedness with Aggregated Time Information}

In order to find how \textit{transient links}, links that do not persist over different times, and \textit{stable links}, links that persist over time, affect the relatedness of the entities, we constructed different models of aggregated graphs as described in Section~\ref{sec:aggregated-model}. We then applied variations of baseline Jaccard methods described previously and the proposed extended Jaccard similarity with reciprocal centralities over the models. 

		\begin{table*}[!h]
		\centering
		\captionof{table}{The comparison of Spearman Correlations to the KORE gold standard of different methods from each dataset over time-varying graphs and aggregated graphs with redirects}
		\label{tab:corr-all}
		\begin{threeparttable}
			\begin{tabular}{|l|l|l|l|l|l|l|l|l|l|}
				\hline
				& Jaccard (I+O)  & \begin{tabular}[c]{@{}l@{}} Extended Jaccard RP \\ (I+O) \end{tabular}  & \begin{tabular}[c]{@{}l@{}} Extended Jaccard RD \\ (I+O) \end{tabular}  \\ \hline
				2007         & 0.522472 $***$ & 0.530393 $***$            & 0.540804 $**$             \\ \hline
				2008         & 0.532665 $**$  & 0.529899 $***$            & 0.580926 $**$             \\ \hline
				2009         & 0.580944 $**$  & 0.613450 $*$              & 0.651450 $\star$              \\ \hline
				2010         & 0.586593 $***$ & 0.637930 $**$             & 0.696506                  \\ \hline
				2011         & 0.561152 $**$  & 0.619950 $*$              & 0.669867                  \\ \hline
				2012         & 0.537278 $***$ & 0.582106 $**$             & 0.634255 $\star$              \\ \hline
				2013         & 0.543986 $***$ & 0.589968 $**$             & 0.640649 $\star$              \\ \hline
				2014         & 0.483211 $***$ & 0.530217 $***$            & 0.589831 $*$              \\ \hline
				2015         & 0.506170 $***$ & 0.529598 $***$            & 0.579989 $**$             \\ \hline
				2016         & 0.494577 $***$ & 0.520003 $***$            & 0.544250 $**$             \\ \hline
				Intersection & 0.470378 $***$ & 0.465568 $***$            & 0.487342 $***$            \\ \hline
				Union        & 0.531830 $***$ & 0.652417 $***$            & \textbf{0.708271}         \\ \hline
			\end{tabular}
			\begin{tablenotes}[para,flushleft]
				$\star$, $\ast$, $\ast\ast$, and $\ast\ast\ast$ mean the Union model using the Extended Jaccard with Reciprocal Degree Centrality considering both in-links and out-links is significantly better than this result with p-value $<$ 0.1, p-value $<$ 0.05, p-value $<$ 0.01 and p-value $<$ 0.001 respectively.
			\end{tablenotes}
		\end{threeparttable}
		\end{table*}

Table~\ref{tab:corr-all} shows the comparison of Spearman Correlations to the KORE gold standard of different methods from each dataset over time-varying graphs and aggregated graphs with redirects. $\star$,$\ast$,$\ast\ast$ and $\ast\ast\ast$ show that the results are significantly lower than the union model with the Extended Jaccard with Reciprocal Degree Centrality considering both in-links and out-links  where p-value $<$ 0.1, p-value $<$ 0.05, p-value $<$ 0.01 and p-value $<$ 0.001 respectively. I+O for both in-links, RP for reciprocal PageRank and RD for reciprocal degree centrality. We can see that aggregating temporal information as a union graph gives more relatedness accuracy than the results from each dataset from time-varying graphs. Intersection graph gives the least relatedness accuracy to KORE dataset but it represents entities that are strongly related to each other at all time. The examples of the top 3 entities with the highest relatedness scores from the Intersection model with the Extended Jaccard Similarity with Reciprocal Degree Centrality considering both in-links and out-links of \textit{Apple\_Inc.} are \textit{Macbook} \textit{IPhone} and \textit{Apple\_Store\_(online)}.

We also compared our results with the recent graph-based approaches by Hulpus et.al.~\cite{hulpus_path-based_2015} which used the path-based relatedness measures with DBpedia 2014 version\footnote{\url{http://wiki.dbpedia.org/Downloads2014}}, and Freebase\footnote{\url{https://developers.google.com/freebase}} dump from 18 January 2015 as knowledge bases without considering temporal information. The best Spearman correlation results to the KORE gold standard of the approaches are \textit{0.63} using DBpedia, and \textit{0.64} using Freebase as knowledge bases. We can see that our approaches outperforms the recent approaches.

We conducted the experiment to gradually accumulate the dataset starting from DBpedia 2007 increasing one dataset to DBpedia 2016 to find how temporal information affects entity relatedness. We found that the correlations significantly increase when each dataset is added and become stable after DBpedia 2011 as shown in Table~\ref{tab:union-acc}. $\star$ and $\ast$ show that the results are significantly lower than when the following year is added, using the Extended Jaccard with Reciprocal Degree Centrality considering both in-links and out-links, where p-value $<$ 0.1 and p-value $<$ 0.05 respectively.

\begin{table*}[]
\centering
\caption{Spearman Correlations to the KORE gold standard comparing the union models accumulate from DBpedia 2007 to DBpedia 2016}
\label{tab:union-acc}
\begin{threeparttable}
\begin{tabular}{|l|l|}
\hline
Dataset                & Extended Jaccard with Reciprocal Degree Centrality (I+O) \\ \hline
DBpedia 2007    & 0.540804 $\star$                                         \\ \hline
Union 2007-2008 & 0.611253 $*$                                             \\ \hline
Union 2007-2009 & 0.655040 $\star$                                         \\ \hline
Union 2007-2010 & 0.677602 $*$                                             \\ \hline
Union 2007-2011 & 0.700322                                                 \\ \hline
Union 2007-2012 & 0.704762                                                 \\ \hline
Union 2007-2013 & 0.709273                                                 \\ \hline
Union 2007-2014 & 0.706982                                                 \\ \hline
Union 2007-2015 & 0.704117                                                 \\ \hline
Union 2007-2016 & 0.708271                                                 \\ \hline
\end{tabular}
\begin{tablenotes}[para,flushleft]
	$\star$ and $\ast$ mean the results are significantly lower than when the following year is added, using the Extended Jaccard with Reciprocal Degree Centrality considering both in-links and out-links, where p-value $<$ 0.1 and p-value $<$ 0.05 respectively.
\end{tablenotes}
\end{threeparttable}
\end{table*}

When we use temporal information with reciprocal centralities as weights of the extended Jaccard method, the relatedness accuracy become lower as information from after the creation of the KORE dataset is taken into account. Table~\ref{tab:ext-tw-redirects} shows the comparison of Spearman Correlations to the KORE gold standard of different methods over different models of aggregated graphs with redirects. We can see that the Union model with exponential temporal weight gives the lowest accuracy as the most updated information get the highest influence in the calculation.

\begin{table*}[h]
	\centering
	\caption{The comparison of Spearman Correlations to the KORE gold standard of different methods over different models of aggregated graphs with redirects}
	\label{tab:ext-tw-redirects}
	\begin{tabular}{|l|l|l|l|l|l|l|}
		\hline
		\multirow{2}{*}{}           & \multicolumn{3}{l|}{Extended Jaccard TWxRP} & \multicolumn{3}{l|}{Extended Jaccard TWxRD} \\ \cline{2-7} 
		& I             & O            & I+O          & I             & O            & I+O          \\ \hline
		Uniform weight              & 0.6420      & 0.6366     & 0.6481     & 0.6973      & 0.6322     & 0.7008     \\ \hline
		Linear temporal weight      & 0.6400      & 0.6351     & 0.6429     & 0.6925      & 0.6341     & 0.6944     \\ \hline
		Exponential temporal weight  & 0.5603      & 0.6138     & 0.5404     & 0.5446      & 0.5854     & 0.5504     \\ \hline
	\end{tabular}

\end{table*}

\subsection{Frequency of the Relatedness Changes} 

As in the previous section, we shown that the proposed extended Jaccard similarity with reciprocal centralities outperforms text-based similarity and graph-based binary Jaccard similarity in term of relatedness score accuracy. In this section, we show how relatedness changes in twice-a-month time steps using the extended Jaccard similarity with reciprocal degree centrality which gives the best relatedness result. Twice-a-month time steps are used instead of biweekly time steps as the Wikipedia dumps are provided twice a month on the first and the twentieth of each month.

For each Wikipedia dumps from 20 August 2016 to 1 January 2017 as described in Section~\ref{sec:dataset}, we constructed the network of article links seeding the entities from the KORE dataset. Then, we applied the extended Jaccard similarity with reciprocal degree centrality described previously.

Table~\ref{tab:corr-diff-wiki} shows the correlations of the entity relatedness scores between each Wikipedia dataset, for example, 2016-08-20 is a Wikipedia 2016-08-20 dataset which is a dump from 20 August 2016. We can see that the relatedness scores are highly correlated within twice-a-month time steps. Although, it is slightly different when time pass, the differences are not statistically significant. Hence, using DBpedia can give enough information for overall relatedness changes which also has more historical data available.

\begin{table*}[]
		\centering
		\captionof{table}{Correlations of the entity relatedness scores between each Wikipedia dataset with redirects}
\label{tab:corr-diff-wiki}
    \resizebox{\textwidth}{!}{
\begin{tabular}{|l|l|l|l|l|l|l|l|l|l|}
\hline
\textbf{Dataset}    & \textbf{2016-09-01} & \textbf{2016-09-20} & \textbf{2016-10-01} & \textbf{2016-10-20} & \textbf{2016-11-01} & \textbf{2016-11-20} & \textbf{2016-12-01} & \textbf{2016-12-20} & \textbf{2017-01-01} \\ \hline
\textbf{2016-08-20} & 0.985076            & 0.984246            & 0.984019            & 0.983054            & 0.964944            & 0.964642            & 0.962263            & 0.962037            & 0.952581            \\ \hline
\textbf{2016-09-01} &                     & 0.999170            & 0.998943            & 0.997978            & 0.979867            & 0.979566            & 0.977187            & 0.976583            & 0.967127            \\ \hline
\textbf{2016-09-20} &                     &                     & 0.999623            & 0.998658            & 0.980849            & 0.980397            & 0.977905            & 0.977753            & 0.968257            \\ \hline
\textbf{2016-10-01} &                     &                     &                     & 0.999036            & 0.980850            & 0.980398            & 0.978207            & 0.977904            & 0.968258            \\ \hline
\textbf{2016-10-20} &                     &                     &                     &                     & 0.981181            & 0.980729            & 0.978237            & 0.978085            & 0.968590            \\ \hline
\textbf{2016-11-01} &                     &                     &                     &                     &                     & 0.999118            & 0.996819            & 0.996457            & 0.987050            \\ \hline
\textbf{2016-11-20} &                     &                     &                     &                     &                     &                     & 0.997408            & 0.996908            & 0.987352            \\ \hline
\textbf{2016-12-01} &                     &                     &                     &                     &                     &                     &                     & 0.998630            & 0.988951            \\ \hline
\textbf{2016-12-20} &                     &                     &                     &                     &                     &                     &                     &                     & 0.990171            \\ \hline
\end{tabular}}
\end{table*}



\section{Summary and Discussions}
We have addressed the research questions of how to represent Wikipedia article link network and how to find relatedness and its evolution using the network.
We have seen that using the redirects and both in and out links provides the best representation of the Wikipedia link network and that the union of these provides the best temporal representation. We have also introduced a new extended Jaccard method that can effectively capture the relatedness and evolution of entities in Wikipedia.

The proposed graph-based extended Jaccard similarity with reciprocal centralities outperform the baseline text-based approach and graph-based Jaccard methods. Moreover, adding more semantic relation using redirect relationships into the models improve accuracy of the relatedness scores. The relatedness score from DBpedia 2010, which is the closest time period to when the KORE dataset was created, are the most correlated to KORE dataset and using the most recent version of DBpedia loses a lot of accuracy. This shows that even in a short space of time the evaluations made by the annotators of the KORE dataset have become outdated. However, we show that by aggregating temporal information as one graph, the accuracy of relatedness is better than any other results from each dataset, as well as the recent research, demonstrating the value of considering not just the most recent version of a semantic graph but also temporal information when performing entity relatedness.
Furthermore, our method scales well with the number of entity pairs, as the cost to estimate similarity of a pair is constant, thus our approach is linear with the size of dataset, with the main cost being the construction of the 2-hop article link network.  

We have seen that the entity relatedness varies according to the relatedness of the entities at that time. With twice-a-month time steps, there is no statistically significant difference in relatedness changes. 
However, with yearly time steps 
accumulated changes in the underlying information lead to significant changes in the relatedness. However, changes will be detected with a delay caused by the distance between time steps. With more fine grain time steps, this can be used to detect emerging events if entities that are not related before become closely related.

A limitation of our approach of using Wikipedia dumps is that if the data is captured at the time that the articles was vandalized, the information may be distorted. This can be improve by using vandalism detection to eliminate the article versions which are detected as vandalism and choose the previous article versions that are not vandalism instead.

As future work, this work could be improved by applying disambiguation techniques to get more entities in additional to user-provided Wikipedia links to add more information to the networks. This work could also be used in a use case such as recommender systems, especially for time-sensitive recommendations.

\section*{Acknowledgements} 
This work was supported by Science Foundation Ireland (SFI) under Grant Numbers SFI/12/RC/2289 (Insight).

\bibliographystyle{acl_natbib}
\bibliography{main}

\begin{thebibliography}{26}
\expandafter\ifx\csname natexlab\endcsname\relax\def\natexlab#1{#1}\fi

\bibitem[{Aggarwal and Buitelaar(2014)}]{aggarwal_wikipedia-based_2014}
Nitish Aggarwal and Paul Buitelaar. 2014.
\newblock {Wikipedia-based Distributional Semantics for Entity Relatedness}.
\newblock In \emph{Association for the Advancement of Artificial Intelligence
  (AAAI) Fall Symposium}, AAAI-2014.

\bibitem[{Althoff et~al.(2015)Althoff, Dong, Murphy, Alai, Dang, and
  Zhang}]{althoff_timemachine:_2015}
Tim Althoff, Xin~Luna Dong, Kevin Murphy, Safa Alai, Van Dang, and Wei Zhang.
  2015.
\newblock Timemachine: Timeline generation for knowledge-base entities.
\newblock In \emph{Proceedings of the 21th ACM SIGKDD International Conference
  on Knowledge Discovery and Data Mining}, KDD '15, pages 19--28, New York, NY,
  USA. ACM.

\bibitem[{Auer et~al.(2007)Auer, Bizer, Kobilarov, Lehmann, Cyganiak, and
  Ives}]{auer2007dbpedia}
S{\"o}ren Auer, Christian Bizer, Georgi Kobilarov, Jens Lehmann, Richard
  Cyganiak, and Zachary Ives. 2007.
\newblock {DBpedia}: A nucleus for a web of open data.
\newblock \emph{The Semantic Web}, pages 722--735.

\bibitem[{Auer and Herre(2006)}]{auer2006versioning}
S{\"o}ren Auer and Heinrich Herre. 2006.
\newblock A versioning and evolution framework for {RDF} knowledge bases.
\newblock In \emph{International Andrei Ershov Memorial Conference on
  Perspectives of System Informatics}, pages 55--69. Springer.

\bibitem[{Bairi et~al.(2015)Bairi, Carman, and
  Ramakrishnan}]{bairi_evolution_2015}
Ramakrishna~B. Bairi, Mark Carman, and Ganesh Ramakrishnan. 2015.
\newblock On the {Evolution} of {Wikipedia}: {Dynamics} of {Categories} and
  {Articles}.
\newblock In \emph{Ninth {International} {AAAI} {Conference} on {Web} and
  {Social} {Media}}.

\bibitem[{Borra et~al.(2015)Borra, Weltevrede, Ciuccarelli, Kaltenbrunner,
  Laniado, Magni, Mauri, Rogers, and Venturini}]{borra_2015_societal}
Erik Borra, Esther Weltevrede, Paolo Ciuccarelli, Andreas Kaltenbrunner, David
  Laniado, Giovanni Magni, Michele Mauri, Richard Rogers, and Tommaso
  Venturini. 2015.
\newblock Societal controversies in {W}ikipedia articles.
\newblock In \emph{Proceedings of the 33rd Annual ACM Conference on Human
  Factors in Computing Systems}, CHI '15, pages 193--196, New York, NY, USA.
  ACM.

\bibitem[{Ceroni et~al.(2014)Ceroni, Georgescu, Gadiraju, Naini, and
  Fisichella}]{ceroni_information_2014}
Andrea Ceroni, Mihai Georgescu, Ujwal Gadiraju, Kaweh~Djafari Naini, and Marco
  Fisichella. 2014.
\newblock Information {Evolution} in {Wikipedia}.
\newblock In \emph{Proceedings of {The} {International} {Symposium} on {Open}
  {Collaboration}}, {OpenSym} '14, pages 24:1--24:10, New York, NY, USA. ACM.

\bibitem[{Das~Sarma et~al.(2011)Das~Sarma, Jain, and
  Yu}]{das_sarma_dynamic_2011}
Anish Das~Sarma, Alpa Jain, and Cong Yu. 2011.
\newblock Dynamic {Relationship} and {Event} {Discovery}.
\newblock In \emph{Proceedings of the {Fourth} {ACM} {International}
  {Conference} on {Web} {Search} and {Data} {Mining}}, {WSDM} '11, pages
  207--216, New York, NY, USA. ACM.

\bibitem[{Gabrilovich and Markovitch(2007)}]{gabrilovich_computing_2007}
Evgeniy Gabrilovich and Shaul Markovitch. 2007.
\newblock Computing {Semantic} {Relatedness} {Using} {Wikipedia}-based
  {Explicit} {Semantic} {Analysis}.
\newblock In \emph{Proceedings of the 20th {International} {Joint} {Conference}
  on {Artifical} {Intelligence}}, {IJCAI}'07, pages 1606--1611, San Francisco,
  CA, USA. Morgan Kaufmann Publishers Inc.

\bibitem[{Hienert and Luciano(2015)}]{hienert_extraction_2012}
Daniel Hienert and Francesco Luciano. 2015.
\newblock Extraction of historical events from {W}ikipedia.
\newblock In \emph{The Semantic Web: ESWC 2012 Satellite Events: ESWC 2012
  Satellite Events, Heraklion, Crete, Greece, May 27-31, 2012. Revised Selected
  Papers}, pages 16--28, Berlin, Heidelberg. Springer Berlin Heidelberg.

\bibitem[{Hoffart et~al.(2012)Hoffart, Seufert, Nguyen, Theobald, and
  Weikum}]{hoffart_kore_2012}
Johannes Hoffart, Stephan Seufert, Dat~Ba Nguyen, Martin Theobald, and Gerhard
  Weikum. 2012.
\newblock Kore: Keyphrase overlap relatedness for entity disambiguation.
\newblock In \emph{Proceedings of the 21st ACM International Conference on
  Information and Knowledge Management}, CIKM '12, pages 545--554, New York,
  NY, USA. ACM.

\bibitem[{Hoffart et~al.(2013)Hoffart, Suchanek, Berberich, and
  Weikum}]{hoffart_yago2_2013}
Johannes Hoffart, Fabian~M. Suchanek, Klaus Berberich, and Gerhard Weikum.
  2013.
\newblock {{YAGO2}: A Spatially and Temporally Enhanced Knowledge Base from
  {Wikipedia}}.
\newblock \emph{Artificial Intelligence}, 194:28 -- 61.

\bibitem[{Hulpus et~al.(2015)Hulpus, Prangnawarat, and
  Hayes}]{hulpus_path-based_2015}
Ioana Hulpus, Narumol Prangnawarat, and Conor Hayes. 2015.
\newblock Path-{Based} {Semantic} {Relatedness} on {Linked} {Data} and {Its}
  {Use} to {Word} and {Entity} {Disambiguation}.
\newblock In \emph{The {Semantic} {Web} - {ISWC} 2015}, pages 442--457.
  Springer, Cham.

\bibitem[{Leal et~al.(2012)Leal, Rodrigues, and
  Queir{\'o}s}]{leal_computing_2012}
Jos{\'e}~Paulo Leal, V{\^a}nia Rodrigues, and Ricardo Queir{\'o}s. 2012.
\newblock {Computing Semantic Relatedness using DBPedia}.
\newblock In \emph{1st Symposium on Languages, Applications and Technologies},
  volume~21 of \emph{OpenAccess Series in Informatics (OASIcs)}, pages
  133--147, Dagstuhl, Germany. Schloss Dagstuhl--Leibniz-Zentrum fuer
  Informatik.

\bibitem[{Lehmann et~al.(2015)Lehmann, Isele, Jakob, Jentzsch, Kontokostas,
  Mendes, Hellmann, Morsey, Van~Kleef, Auer et~al.}]{lehmann2015dbpedia}
Jens Lehmann, Robert Isele, Max Jakob, Anja Jentzsch, Dimitris Kontokostas,
  Pablo~N Mendes, Sebastian Hellmann, Mohamed Morsey, Patrick Van~Kleef,
  S{\"o}ren Auer, et~al. 2015.
\newblock {DBpedia}--a large-scale, multilingual knowledge base extracted from
  {Wikipedia}.
\newblock \emph{Semantic Web}, 6(2):167--195.

\bibitem[{Nunes et~al.(2008)Nunes, Ribeiro, and
  David}]{nunes_wikichanges:_2008}
Sérgio Nunes, Cristina Ribeiro, and Gabriel David. 2008.
\newblock {WikiChanges}: {Exposing} {Wikipedia} {Revision} {Activity}.
\newblock In \emph{Proceedings of the 4th {International} {Symposium} on
  {Wikis}}, {WikiSym} '08, pages 25:1--25:4, New York, NY, USA. ACM.

\bibitem[{Page et~al.(1999)Page, Brin, Motwani, and
  Winograd}]{page_pagerank_1999}
Lawrence Page, Sergey Brin, Rajeev Motwani, and Terry Winograd. 1999.
\newblock The {P}age{R}ank citation ranking: Bringing order to the {W}eb.
\newblock Technical Report 1999-66, Stanford InfoLab.

\bibitem[{Prangnawarat and Hayes(2017)}]{prangnawarat_temporal_2017}
Narumol Prangnawarat and Conor Hayes. 2017.
\newblock Temporal evolution of entity relatedness using {Wikipedia} and
  {DBpedia}.
\newblock In \emph{Workshop on Managing the Evolution and Preservation of the
  Data Web (MEPDaW 2017)}.

\bibitem[{Radinsky et~al.(2011)Radinsky, Agichtein, Gabrilovich, and
  Markovitch}]{radinsky_word_2011}
Kira Radinsky, Eugene Agichtein, Evgeniy Gabrilovich, and Shaul Markovitch.
  2011.
\newblock A {Word} at a {Time}: {Computing} {Word} {Relatedness} {Using}
  {Temporal} {Semantic} {Analysis}.
\newblock In \emph{Proceedings of the 20th {International} {Conference} on
  {World} {Wide} {Web}}, {WWW} '11, pages 337--346, New York, NY, USA. ACM.

\bibitem[{Strehl et~al.(2000)Strehl, Ghosh, and Mooney}]{strehl2000impact}
Alexander Strehl, Joydeep Ghosh, and Raymond Mooney. 2000.
\newblock Impact of similarity measures on {W}eb-page clustering.
\newblock In \emph{Workshop on Artificial Intelligence for Web Search (AAAI
  2000)}, volume~58, page~64.

\bibitem[{Strube and Ponzetto(2006)}]{strube_wikirelate!_2006}
Michael Strube and Simone~Paolo Ponzetto. 2006.
\newblock {WikiRelate}! {Computing} {Semantic} {Relatedness} {Using}
  {Wikipedia}.
\newblock In \emph{Proceedings of the 21st {National} {Conference} on
  {Artificial} {Intelligence} - {Volume} 2}, {AAAI}'06, pages 1419--1424,
  Boston, Massachusetts. AAAI Press.

\bibitem[{Tanimoto(1958)}]{tanimoto_elementary_1958}
T.T. Tanimoto. 1958.
\newblock \emph{An Elementary Mathematical Theory of Classification and
  Prediction}.
\newblock International Business Machines Corporation.

\bibitem[{Tran et~al.(2017)Tran, Tran, and Nieder{\'e}e}]{Tran_Beyond_2017}
Nam~Khanh Tran, Tuan Tran, and Claudia Nieder{\'e}e. 2017.
\newblock \emph{Beyond Time: Dynamic Context-Aware Entity Recommendation}.
  Springer International Publishing.

\bibitem[{Tran et~al.(2014)Tran, Ceroni, Georgescu, Naini, and
  Fisichella}]{tran_wikipevent:_2014}
Tuan Tran, Andrea Ceroni, Mihai Georgescu, Kaweh~Djafari Naini, and Marco
  Fisichella. 2014.
\newblock {WikipEvent}: {Leveraging} {Wikipedia} {Edit} {History} for {Event}
  {Detection}.
\newblock In \emph{Web {Information} {Systems} {Engineering} – {WISE} 2014},
  Lecture {Notes} in {Computer} {Science}, pages 90--108. Springer, Cham.

\bibitem[{V{\"o}lkel and Groza(2006)}]{volkel2006semversion}
Max V{\"o}lkel and Tudor Groza. 2006.
\newblock {SemVersion}: An {RDF}-based ontology versioning system.
\newblock In \emph{Proceedings of the IADIS international conference
  WWW/Internet}, volume 2006, page~44.

\bibitem[{Whiting et~al.(2014)Whiting, Jose, and
  Alonso}]{whiting_wikipedia_2014}
Stewart Whiting, Joemon Jose, and Omar Alonso. 2014.
\newblock Wikipedia {As} a {Time} {Machine}.
\newblock In \emph{Proceedings of the 23rd {International} {Conference} on
  {World} {Wide} {Web}}, {WWW} '14 {Companion}, pages 857--862, New York, NY,
  USA. ACM.

\end{thebibliography}

\end{document}